\documentclass[runningheads]{llncs}

\usepackage{graphicx}
\usepackage{comment}
\usepackage{amsmath,amssymb}
\usepackage{color}
\usepackage{url}
\usepackage{hyperref}

\usepackage{subcaption}
\usepackage{tabularx}
\usepackage{array,multirow}
 \usepackage{booktabs}
 \usepackage[dvipsnames]{xcolor}
\usepackage{arydshln}
\usepackage{caption}
\usepackage{array,multirow}
\newif\ifreview
\reviewfalse

\ifreview
	\usepackage{lineno}

	\linenumbers
\fi

\begin{document}


\def\SubNumber{38}

\def\GCPRTrack{Main Track}

\title{Rethinking Semi-supervised Segmentation Beyond Accuracy: Reliability and Robustness}

\ifreview
\else
	\titlerunning{Rethinking Semi-supervised Segmentation Beyond Accuracy}


	\author{Steven Landgraf, Markus Hillemann, Markus Ulrich}
	
	\authorrunning{S. Landgraf et al.}
	
	\institute{Institute of Photogrammetry and Remote Sensing (IPF), \\
Karlsruhe Institute of Technology (KIT), Karlsruhe, Germany. \\
	\email{\{steven.landgraf, markus.hillemann, markus.ulrich\}@kit.edu}}
\fi

\maketitle              

\begin{abstract}
Semantic segmentation is critical for scene understanding but demands costly pixel-wise annotations, attracting increasing attention to semi-supervised approaches to leverage abundant unlabeled data. While semi-supervised segmentation is often promoted as a path toward scalable, real-world deployment, it is astonishing that current evaluation protocols exclusively focus on segmentation accuracy, entirely overlooking reliability and robustness. These qualities, which ensure consistent performance under diverse conditions (robustness) and well-calibrated model confidences as well as meaningful uncertainties (reliability), are essential for safety-critical applications like autonomous driving, where models must handle unpredictable environments and avoid sudden failures at all costs. To address this gap, we introduce the Reliable Segmentation Score (RSS), a novel metric that combines predictive accuracy, calibration, and uncertainty quality measures via a harmonic mean. RSS penalizes deficiencies in any of its components, providing an easy and intuitive way of holistically judging segmentation models. Comprehensive evaluations of UniMatchV2 against its predecessor and a supervised baseline show that semi-supervised methods often trade reliability for accuracy. While out-of-domain evaluations demonstrate UniMatchV2's robustness, they further expose persistent reliability shortcomings. We advocate for a shift in evaluation protocols toward more holistic metrics like RSS to better align semi-supervised learning research with real-world deployment needs. 

\end{abstract}

\section{Introduction}
Semantic segmentation is one of the most fundamental tasks in scene understanding through pixel-wise classifications \cite{long2015fully,chen2017deeplab,zhao2017pyramid,minaee2021image}. However, to fully utilize the capabilities of modern neural networks, a large quantity of dense annotations is required, which are tremendously costly. On Cityscapes \cite{cordts2016cityscapes}, for instance, a single image with just 19 classes takes around 1.5~hours to label. Needless to say, this is a significant obstacle, hindering real-world deployment.

For this reason, semi-supervised semantic segmentation has garnered increasing attention \cite{pelaez2023survey}. The general idea lies in leveraging large amounts of unlabeled images, accompanied by a handful of manually labeled images to alleviate the massive labeling cost of fully supervised learning \cite{papandreou2015weakly,yang2022survey}. Recent works \cite{yang2022st++,wang2022semi,liu2022perturbed,jin2022semi,xu2022semi,zhao2023instance,zhao2023augmentation,yang2023revisiting,li2023diverse,ma2023enhanced,liang2023logic,wang2024towards,sun2024corrmatch,wang2024allspark,howlader2024beyond,hoyer2024semivl,yang2025unimatch} have managed to steadily improve segmentation quality on various benchmarks datasets \cite{everingham2015pascal,cordts2016cityscapes,zhou2017scene,lin2014microsoft}. Interestingly, in the most recent state-of-the-art approach, "UniMatchV2: Pushing the Limit of Semi-Supervised Semantic Segmentation", Yang et al. \cite{yang2025unimatch} accurately identify that semi-supervised segmentation methods are becoming increasingly sophisticated, and yet performance gains are very marginal. Based on this observation, they argue that it is necessary to switch to more modern Vision Transformer-based model architectures (e.g., DINOv2) \cite{oquab2024dinov2} and focus on more challenging benchmark datasets. 

While we strongly agree with this notion, we argue to consider reliability as a complementary concept in semi-supervised segmentation, i.e., whether a model's confidence reflects the true likelihood of correctness and its uncertainty estimates align with predictive ambiguities and errors. In addition, we emphasize robustness, which refers to a model's ability to maintain high performance under perturbations, noise, or distribution shifts. Although these have already been introduced as an integral component in evaluating semantic segmentation models \cite{de2023reliability,loiseau2024reliability,croce2024towards,yu2023robust,zhou2019automated,oliveira2018efficient}, astonishingly, they have been overlooked entirely in semi-supervised segmentation. This is, perhaps, a bit surprising since semi-supervised learning builds upon the premise of making deep learning accessible for real-world deployment, where a model's prediction need not just be accurate, but more importantly, also reliable and robust. 

Therefore, we advocate rethinking semi-supervised evaluation protocols to prioritize reliability and robustness alongside accuracy. To this end, we propose a novel evaluation metric and evaluate the current state-of-the-art method to guide future research toward real-world deployment needs. More precisely, our contributions can be summarized as follows: 
\begin{enumerate}
    \item We introduce the Reliable Segmentation Score (RSS), a novel metric combining accuracy, calibration, and uncertainty quality via a harmonic mean to penalize deficiencies, enabling holistic assessment of segmentation models’ reliability and robustness.
    \item Additionally, we provide the first comprehensive evaluation of the current state-of-the-art UniMatchV2 \cite{yang2025unimatch} in terms of reliability on two in-domain scenarios, comparing it against UniMatchV1 \cite{yang2023revisiting} and a supervised baseline. Thereby, we reveal that semi-supervised methods often sacrifice calibration and uncertainty quality for accuracy. 
    \item Finally, we perform extensive out-of-domain evaluations on Rainy Cityscapes \cite{hu2019depth} and Foggy Cityscapes \cite{sakaridis2018semantic}, showcasing that UniMatchV2 may be more robust but exhibits persistent reliability shortcomings.
\end{enumerate}

We firmly believe that taking reliability and robustness into account is crucial to understand whether semi-supervised segmentation research is moving in the originally intended direction or just incrementally improving the accuracy on selected benchmark datasets. 

\section{Related Work}

\subsection{Semi-supervised Semantic Segmentation}
While all semi-supervised semantic segmentation methods try to leverage unlabeled images in addition to a handful of manually labeled images, they can be divided into five categories based on the taxonomy proposed by Peláez et al. \cite{pelaez2023survey}. The first category comprises adversarial methods, which either exploit generative models to create synthetic images to incorporate them into the segmentation task \cite{souly2017semi,li2021semantic},  or GAN-like structures \cite{goodfellow2014generative}, where the segmentation model itself acts as the generator, and a discriminator discerns predicted segmentation maps and the real ground truth labels \cite{hung2018adversarial,mittal2019semi,mendel2020semi,ke2020guided,zhang2021stable,jin2021adversarial}. The second category includes consistency regularization techniques that apply perturbations to unlabeled data and train models to remain invariant to these changes, typically achieved through the introduction of an additional regularization term in the loss function \cite{tarvainen2017mean,french2019semi,olsson2021classmix,chen2021complexmix,zhao2023augmentation,chen2021semi,peng2020deep,liu2022perturbed,wu2023perturbation,yang2023revisiting}. In the third category, pseudo-labeling methods generate pseudo-labels of unlabeled images by utilizing a model pre-trained on labeled data. These approaches either follow a self-training protocol based on high-confidence predictions of a single model \cite{yang2022st++,teh2022gist,zhu2021improving,yuan2021simple,he2021re,sun2024corrmatch,wang2024allspark}, or involve multiple models through mutual-training \cite{feng2022dmt,zhou2022catastrophic,li2023semi,li2023cfcg,na2024switching}. The fourth category applies contrastive learning to structure the feature space by clustering similar samples and separating dissimilar ones. Leveraging the success of self-supervised contrastive learning techniques, such as SimCLR \cite{chen2020simple,chen2020big}, several semi-supervised segmentation methods have been developed to enhance feature representations \cite{liu2021bootstrapping,chen2021exploring,alonso2021semi}. Finally, the last category describes hybrid methods that combine elements from the previous four categories, particularly pseudo-labeling and consistency regularization \cite{qiao2023fuzzy,wang2023hunting,wang2023conflict,ma2023enhanced,liang2023logic,wang2023space,hu2024training,mai2024rankmatch,wang2024towards,yang2023revisiting}.

\subsection{Reliability and Robustness}
\textbf{Reliability.} As mentioned above, reliability encompasses both model calibration, i.e., how well predicted probabilities reflect the true likelihood of correctness, and uncertainty quality, which describes a model's ability to align its confidences with predictive ambiguities and errors. We will go over the importance of distinguishing between both of these in Section \ref{sec: rss}. Even though deep neural networks can have impressive predictive performance, they are known to be overconfident \cite{guo2017calibration,wilson2020bayesian,wang2021rethinking}. In this context, Guo et al. \cite{guo2017calibration} proposed temperature scaling as a simple post-hoc calibration technique, which has remained a widely used baseline due to its simplicity and non-invasiveness compared to alternative approaches \cite{kull2019beyond,naeini2015obtaining,ji2019bin,ding2021local,patra2023calibrating}. While most calibration methods offer an effective way of enhancing reliability in in-domain settings, Ovadia et al. \cite{ovadia2019can} observed that calibration deteriorates significantly under domain shifts. This lack of robustness poses a significant risk in safety-critical contexts like autonomous driving \cite{muhammad2020deep}, where models must maintain their reliability despite unexpected scenarios.

\textbf{Robustness.} A model's robustness is defined by the ability to remain effective under perturbations, noise, or distribution shifts. Importantly, robustness encompasses not only accuracy but also reliable confidence and uncertainty estimates, as outlined previously. Prior work distinguishes between robustness to perturbations (Gaussian noise, blur, occlusions), adversarial attacks (imperceptible perturbations crafted to induce failure) \cite{hendrycks2019benchmarking,kamann2020benchmarking,kamann2021benchmarking,goodfellow2014explaining}, and natural domain shifts (variations in weather, lighting, or geographic contexts) \cite{hendrycks2021natural,pedraza2022really,recht2019imagenet,sakaridis2021acdc,hu2019depth,sakaridis2018semantic,varma2019idd}.

\textbf{Research Gap.} While there are plenty of previous analyses on reliability and robustness in semantic segmentation \cite{arnab2018robustness,kamann2020benchmarking,kamann2021benchmarking,xie2021segformer,zhou2022understanding,de2023reliability,loiseau2024reliability,zhou2019automated} -- highlighting the importance of this topic -- none of these considered semi-supervised learning. Instead of proposing yet another semi-supervised segmentation method with marginal gains in segmentation accuracy, we investigate the reliability and robustness of the current state of the art to answer the following, crucial question:
\begin{center} \textit{"Should we rethink semi-supervised segmentation evaluation protocols\\ to prioritize reliability and robustness alongside accuracy?"} \end{center}

\section{Reliable Segmentation Score}\label{sec: rss}
Current semi-supervised semantic segmentation methods evaluate their models almost exclusively with the mean Intersection over Union (mIoU) \cite{lateef2019survey}
\begin{equation}
    \text{mIoU} = \frac{1}{C} \sum_{c=1}^C \frac{TP_c}{FP_c + FN_c + TP_c}\enspace ,
\end{equation}
where $C$ is the number of classes, $TP$ is the number of true positives, $FP$ is the number of false positives, and $FN$ is the number of false negatives. As depicted, the intersection over union is computed on a per-class basis and then averaged, providing a straightforward metric to quantify the segmentation quality of a model.

The Expected Calibration Error (ECE) \cite{guo2017calibration} is foundational to evaluate the calibration of semantic segmentation models, measuring the average mismatch between a model's confidence and its empirical accuracy across $M$ probability bins:
\begin{equation}
    \text{ECE} = \sum_{m=1}^M \frac{|B_m|}{N} \left| \text{acc}(B_m) - \text{conf}(B_m) \right|\enspace,
\end{equation}
where $|B_m|$ is the number of pixels in the $m$-th bin, $N$ is the total number of pixels, $\text{acc}(B_m)$ is the accuracy of predictions in the $m$-th bin, and $\text{conf}(B_m)$ is the average confidence of predictions in the corresponding bin. ECE is vital in safety-critical applications like autonomous driving, where a lack of reliability, e.g., through overconfidence \cite{wilson2020bayesian,wang2021rethinking}, can have disastrous consequences.

Recent research has also emphasized the importance of quantifying the uncertainty inherent to a model's prediction to enhance the reliability, robustness, and explainability \cite{abdar2021review,landgraf2024efficient,landgraf2024dudes,landgraf2024uncertainty_ce,landgraf2025critical}. To evaluate uncertainty quality, Mukhoti et al. \cite{mukhoti2018evaluating} propose the following two simple and intuitive metrics: 
\begin{enumerate}
    \item \textbf{p(accurate$|$certain):} The probability that the model is accurate given that the uncertainty is below a specified threshold.
    \item \textbf{p(uncertain$|$inaccurate):} The probability that the uncertainty of the model exceeds a specified threshold given that the prediction is inaccurate.
\end{enumerate}
These conditional probabilities can be calculated as
\begin{equation}
\begin{aligned}
    p(\text{acc}|\text{cer}) = \frac{n_{ac}}{(n_{ac} + n_{ic})} \enspace , \\
    p(\text{unc}|\text{inacc}) = \frac{n_{iu}}{(n_{iu} + n_{ic})}\enspace,
\end{aligned}
\end{equation}
where $n_{ac}$ represents the number of pixels that are accurate and certain, $n_{ic}$ the number of pixels that are inaccurate and certain, and $n_{iu}$ the number of pixels that are inaccurate and uncertain. Undoubtedly, the choice of the threshold is significant, with prior work suggesting the median uncertainty of an image as a suitable default \cite{landgraf2025comparative}.

To provide an easy, intuitive, and holistic assessment of semantic segmentation models, the Reliable Segmentation Score (RSS) combines the previous metrics using the harmonic mean:
\begin{equation}\label{eq: rss}
    \text{RSS} = \frac{\sum \omega_i}{\frac{\omega_1}{\text{mIoU}} + \frac{\omega_2}{(1-\text{ECE})} + \frac{\omega_3}{\text{p(acc$|$cer)}} + \frac{\omega_3}{\text{p(unc$|$inacc)}}}\enspace,
\end{equation}
where $\omega_i$ are application-specific weights, e.g. $\omega_1=1.0, \omega_2=\omega_3=\omega_4=\frac{1}{3}$. To avoid any assumptions about application priorities, we propose to use equal weighting to appropriately reflect the importance of reliability metrics in safety-critical applications, where calibration and uncertainty quality may outweigh raw accuracy. 

RSS integrates accuracy (mIoU), calibration (ECE), and uncertainty quality (p(acc$|$cer) and p(unc$|$inacc)) into a single metric, leveraging the harmonic mean to penalize poor performance in any component. This approach ensures that a model cannot achieve a high RSS unless it excels in all aspects -- accurate predictions, well-calibrated confidences, and reliable uncertainty estimates -- providing a strict yet comprehensive measure to evaluate reliability and robustness.

A key question is whether mIoU, ECE, p(acc$|$cer), and p(unc$|$inacc) evaluate distinct aspects of model performance, justifying their combination in RSS. We argue that these metrics are largely orthogonal, each capturing complementary properties:
\begin{itemize}
    \item \textbf{mIoU:} Measures pixel-wise accuracy, reflecting the model's ability to correctly classify pixels across all classes. It does not consider any reliability-related aspects.
    \item \textbf{ECE:} Assesses calibration, ensuring confidence aligns with accuracy. A model can achieve high mIoU but poor ECE if it is overconfident or underconfident, highlighting their independence.
    \item \textbf{p(acc$|$cer):} Evaluates whether low-uncertainty predictions are correct, focusing on the utility of certainty in decision-making. 
    \item \textbf{p(unc$|$inacc):} Assesses whether incorrect predictions are flagged with high uncertainty, enabling error detection or mitigation.
\end{itemize}

Moreover, mIoU and ECE are fundamentally tied to the confidence of a prediction, i.e., the maximum softmax probability. Both mIoU -- through its reliance on hard class decisions -- and ECE -- which only bins max-softmax values -- therefore evaluate only the single highest probability output. This constrained view, which misses whether a model's output is narrowly peaked or spread across multiple classes, is clearly insufficient and was already shown to lead to inconsistencies between different calibration measures \cite{wolf2024decoupling}. In contrast, the conditional uncertainty probabilities p(acc$|$cer) and p(unc$|$inacc) rely on an uncertainty estimate derived from the entire softmax distribution, like the Shannon entropy \cite{shannon1948mathematical}. This distribution-wide view captures the valuable information about ambiguity between top predictions, which is neglected by mIoU and ECE. Consequently, a model can be well-calibrated (low ECE) yet still fail to flag its inaccuracies as uncertain (low p(unc$|$inacc)), or conversely be accurate when certain (high p(acc$|$cer)) but suffer from being overly certain on its errors (low p(unc$|$inacc)). Hence, by pairing max-softmax-based metrics (mIoU, ECE) -- which ensure high accuracy and guard against overconfidence or underconfidence in the predicted class -- with conditional uncertainty metrics that require low entropy for correct predictions (p(acc$|$cer)) and high entropy for errors (p(unc$|$inacc)), RSS ensures a model is both confident when it should be and uncertain whenever it is inaccurate, yielding a balanced, holistic reliability score.


\section{Experimental Setup}

\subsection{Methodological Background} UniMatchV1 \cite{yang2023revisiting} and UniMatchV2 \cite{yang2025unimatch} are recent and representative approaches in semi-supervised semantic segmentation that build on consistency regularization, leveraging unlabeled data by enforcing prediction consistency under image- and feature-level perturbations.

\textbf{UniMatchV1}, introduced in 2023, extends FixMatch \cite{sohn2020fixmatch} by generating a weakly perturbed image to produce pseudo-labels, which supervise predictions on a strongly perturbed version of the same image. In addition, UniMatchV1 incorporates an auxiliary feature perturbation stream, using dropout \cite{hinton2012improving}, and a dual-stream technique, where two strong views learn from one weak view. 

\textbf{UniMatchV2}, presented in 2025, advances its predecessor by adopting Vision Transformer-based architectures (e.g., DINOv2 \cite{oquab2024dinov2}), pre-trained on large datasets, instead of less potent ResNet encoders \cite{he2016deep}. While it maintains the weak-to-strong consistency framework, it unifies the image-level and feature-level augmentations of UniMatchV1 into a single stream and introduces Complementary Dropout, which applies different dropout masks across the two strongly augmented views to encourage feature diversity. This new design reduces training cost and achieves state-of-the-art results on several benchmarks. 

Together, UniMatchV1 and UniMatchV2 provide a comprehensive evaluation framework. V1 represents a robust and widely adopted approach, while V2 pushes the boundaries of semi-supervised learning with cutting-edge architectures. Their shared weak-to-strong consistency framework allows for direct comparisons of how architectural advancements, i.e., Convolutional Neural Networks vs. Vision Transformers, impact reliability and robustness. Overall, they form a complementary baseline for answering the critical research question if current semi-supervised segmentation methods are not only accurate but also robust and reliable. 

\subsection{Datasets}
We use Cityscapes \cite{cordts2016cityscapes} and Pascal VOC2012 \cite{everingham2015pascal} for in-domain semi-supervised segmentation experiments. For out-of-domain evaluations, we use the validation sets of Foggy Cityscapes \cite{sakaridis2018semantic} and Rainy Cityscapes \cite{hu2019depth}, which introduce increasingly challenging perturbations to the original urban scenes to simulate adverse weather conditions. Foggy Cityscapes offers three different versions, which are defined by the attenuation coefficient $\beta$. Higher $\beta$ values result in thicker fog. Similarly, Rainy Cityscapes offers several options with varying amounts of rain and fog. We evaluate on three sets of parameters, where Rainy$_1$ uses [0.01, 0.005, 0.01], Rainy$_2$ uses [0.02, 0.01, 0.005], and Rainy$_3$ uses [0.03, 0.015, 0.002] as attenuation coefficients $\alpha$ and $\beta$ and the raindrop radius $a$. $\alpha$ and $\beta$ determine the degree of simulated rain and fog in the images.

\subsection{Implementation Details}
All training procedures adhere to the original configurations proposed in the respective UniMatch versions \cite{yang2023revisiting,yang2025unimatch}, including model architectures, hyperparameters, and dataset splits. The supervised baseline is implemented using the same configuration as UniMatchV2.

For evaluation, we follow the UniMatchV2 protocol for computing the mean Intersection-over-Union (mIoU) \cite{yang2025unimatch}. Calibration is assessed using the Expected Calibration Error (ECE), implemented via the torchmetrics package \cite{detlefsen2022torchmetrics} with the default 15-bin discretization.

To quantify the uncertainty, we compute the Shannon entropy \cite{shannon1948mathematical}
\begin{equation}\label{eq: entropy}
H(x) = -\sum_{c=1}^C p(\hat{y}_c(x)) \log p(\hat{y}_c(x))\enspace,
\end{equation}
where $p(\hat{y}_c(x))$ denotes the predicted probability for class $c$ for a given input image $x$. These pixel-wise uncertainty estimates are used to compute the uncertainty metrics, p(acc$|$cer) and p(unc$|$inacc), based on the thresholding strategy suggested by Landgraf et al. \cite{landgraf2025comparative}, categorizing pixels as certain or uncertain based on the median uncertainty within each image. 

\section{Experiments}
In the following, we conduct a comprehensive set of experiments for in-domain reliability and out-of-domain robustness. Moreover, we provide qualitative examples for both scenarios to highlight the importance of not only considering segmentation accuracy but also reliability. Due to space constraints, some evaluation results are included in the supplementary material.

\subsection{In-Domain Evaluation}
\begin{table*}[!ht]
\begin{center}
\setlength\extrarowheight{1mm}
\newcolumntype{C}{>{\centering\arraybackslash}p{1.0cm}}
\resizebox{\textwidth}{!}{
\begin{tabular}{l|c|CCCCC|CCCCC}
\toprule
\multirow{2}{*}{} & \multirow{2}{*}{Encoder} 
& \multicolumn{5}{c|}{Cityscapes (Label Fraction)} 
& \multicolumn{5}{c}{Pascal VOC2012 (Labeled Images)} \\
\cmidrule(lr){3-7} \cmidrule(lr){8-12}
& & 1/30 & 1/16 & 1/8 & 1/4 & 1/2 & 92 & 183 & 366 & 732 & 1464 \\ \hline \hline
\multicolumn{12}{c}{mIoU $\uparrow$} \\ \hline \hline
Supervised & ViT-S     & 0.736 & 0.782 & 0.803 & 0.819 & 0.824 & 0.661 & 0.757 & 0.812 & 0.832 & 0.859 \\
UniMatchV2 & ViT-S     & 0.793 & 0.809 & 0.814 & 0.823 & 0.825 & 0.758 & 0.849 & 0.860 & 0.871 & 0.876 \\
\cdashline{1-12}
Supervised & ViT-B     & 0.788 & 0.811 & 0.829 & 0.840 & 0.846 & 0.765 & 0.824 & 0.854 & 0.876 & 0.886 \\
UniMatchV1 & RN-101       & 0.727 & 0.762 & 0.782 & 0.793 & 0.797 & 0.746 & 0.777 & 0.788 & 0.800 & 0.803 \\
UniMatchV2 & ViT-B     & \textbf{0.807} & \textbf{0.839} & \textbf{0.840} & \textbf{0.847} & \textbf{0.848} & \textbf{0.862} & \textbf{0.873} & \textbf{0.892} & \textbf{0.900} & \textbf{0.903} \\ \hline \hline
\multicolumn{12}{c}{ECE $\downarrow$} \\ \hline \hline
Supervised & ViT-S     & \textbf{0.016} & \textbf{0.023} & \textbf{0.016} & \textbf{0.013} & 0.036 & 0.054 & 0.035 & 0.026 & 0.024 & 0.019 \\
UniMatchV2 & ViT-S     & 0.029 & 0.027 & 0.022 & 0.018 & \textbf{0.011} & 0.053 & 0.028 & 0.025 & 0.021 & 0.019 \\
\cdashline{1-12}
Supervised & ViT-B     & 0.020 & 0.025 & 0.018 & 0.015 & \textbf{0.011} & 0.038 & 0.026 & \textbf{0.019} &\textbf{ 0.018} & 0.017 \\
UniMatchV1 & RN-101       & 0.049 & 0.042 & 0.041 & 0.037 & 0.029 & 0.051 & 0.040 & 0.036 & 0.032 & 0.029 \\
UniMatchV2 & ViT-B     & 0.019 & 0.025 & 0.022 & 0.015 & \textbf{0.011} & \textbf{0.030} & \textbf{0.022} & 0.020 & \textbf{0.018} & \textbf{0.016} \\ \hline \hline
\multicolumn{12}{c}{p(acc$|$cer) $\uparrow$} \\ \hline \hline
Supervised & ViT-S     & 0.880 & 0.900 & 0.876 & 0.880 & \textbf{0.907} & 0.969 & 0.979 & 0.985 & 0.986 & 0.988 \\
UniMatchV2 & ViT-S     & 0.872 & 0.871 & 0.862 & 0.865 & 0.854 & 0.960 & 0.982 & 0.985 & 0.988 & \textbf{0.989} \\
\cdashline{1-12}
Supervised & ViT-B     & \textbf{0.926} & \textbf{0.942} & \textbf{0.942} & \textbf{0.908} & 0.905 & 0.979 & \textbf{0.984} & \textbf{0.989} & \textbf{0.989} & \textbf{0.989} \\
UniMatchV1 & RN-101       & 0.882 & 0.883 & 0.923 & 0.900 & 0.889 & 0.958 & 0.964 & 0.970 & 0.974 & 0.977 \\
UniMatchV2 & ViT-B     & 0.890 & 0.870 & 0.873 & 0.857 & 0.852 & \textbf{0.980} & 0.981 & 0.987 & 0.988 & \textbf{0.989} \\ \hline \hline
\multicolumn{12}{c}{p(unc$|$inacc) $\uparrow$} \\ \hline \hline
Supervised & ViT-S     & 0.663 & 0.720 & 0.641 & 0.648 & \textbf{0.713} & 0.953 & 0.964 & 0.972 & 0.976 & 0.979 \\
UniMatchV2 & ViT-S     & 0.606 & 0.606 & 0.565 & 0.576 & 0.530 & 0.914 & 0.960 & 0.968 & 0.977 & 0.978 \\
\cdashline{1-12}
Supervised & ViT-B     & \textbf{0.794} & \textbf{0.842} & \textbf{0.835} & \textbf{0.709} & 0.698 & \textbf{0.959} & \textbf{0.973} & \textbf{0.979} & \textbf{0.981} & \textbf{0.983} \\
UniMatchV1 & RN-101       & 0.667 & 0.666 & 0.778 & 0.703 & 0.675 & 0.899 & 0.904 & 0.913 & 0.928 & 0.941 \\
UniMatchV2 & ViT-B     & 0.673 & 0.586 & 0.586 & 0.533 & 0.515 & 0.954 & 0.945 & 0.961 & 0.972 & 0.977 \\ \hline \hline
\multicolumn{12}{c}{RSS $\uparrow$} \\ \hline \hline
\text{Supervised} & \text{ViT-S} & 0.797 & 0.833 & 0.806 & 0.814 & 0.841 & 0.860 & 0.906 & 0.930 & 0.938 & 0.949 \\
\text{UniMatchV2} & \text{ViT-S} & 0.786 & 0.790 & 0.772 & 0.780 & 0.758 & 0.887 & 0.938 & 0.944 & 0.951 & 0.953 \\
\cdashline{1-12}
\text{Supervised} & \text{ViT-B} & \textbf{0.864} & \textbf{0.888} & \textbf{0.892} & \textbf{0.848} & \textbf{0.845} & 0.907 & 0.934 & 0.947 & 0.954 & 0.958 \\
\text{UniMatchV1} & \text{RN-101} & 0.791 & 0.802 & 0.853 & 0.828 & 0.818 & 0.879 & 0.894 & 0.902 & 0.912 & 0.917 \\
\text{UniMatchV2} & \text{ViT-B} & 0.822 & 0.789 & 0.790 & 0.763 & 0.754 & \textbf{0.939} & \textbf{0.942} & \textbf{0.953} & \textbf{0.959} & \textbf{0.962} \\ \bottomrule
\end{tabular}}
\end{center}
\caption{In-domain results of a supervised baseline, UniMatchV1 \cite{yang2023revisiting}, and UniMatchV2 \cite{yang2025unimatch} on Cityscapes \cite{cordts2016cityscapes} and Pascal VOC2012 \cite{everingham2015pascal} across different training splits. The quantitative comparison includes two DINOv2 encoders \cite{oquab2024dinov2} -- ViT-S and ViT-B -- as well as the conventional ResNet-101 (RN-101) backbone \cite{he2016deep}. Best results are marked in \textbf{bold}.}
\label{table: id_quantitative_comparison}
\end{table*}

\textbf{Quantitative Analysis.} Table \ref{table: id_quantitative_comparison} summarizes in-domain results on Cityscapes and Pascal VOC2012 across different label regimes. As expected, UniMatchV2 consistently outperforms both the supervised baseline and UniMatchV1 in mIoU, especially with fewer available labels, and segmentation quality generally improves with the larger encoder and more labeled images. However, the supervised baseline excels in terms of reliability, outperforming both semi-supervised approaches concerning ECE, p(acc$|$cer), and p(unc$|$inacc) in most cases, particularly on Cityscapes. This is highlighted by our proposed RSS metric, where the supervised ViT-B baseline achieves the highest results across all splits on Cityscapes. Interestingly, even UniMatchV1 attains a higher RSS in 4 out of the 5 splits compared to UniMatchV2, primarily due to the inability of UniMatchV2 to assign high uncertainties to incorrect predictions, as measured by p(unc$|$inacc). Nonetheless, UniMatchV2 still offers the highest RSS across all label regimes on Pascal VOC2012 due to its remarkable improvements over the supervised baseline and UniMatchV1 concerning segmentation quality as well as staying competitive in terms of the three reliability metrics. 

\begin{figure*}[!ht]
    \centering
    
    \begin{subfigure}{0.99\textwidth}
        \includegraphics[width=\textwidth]{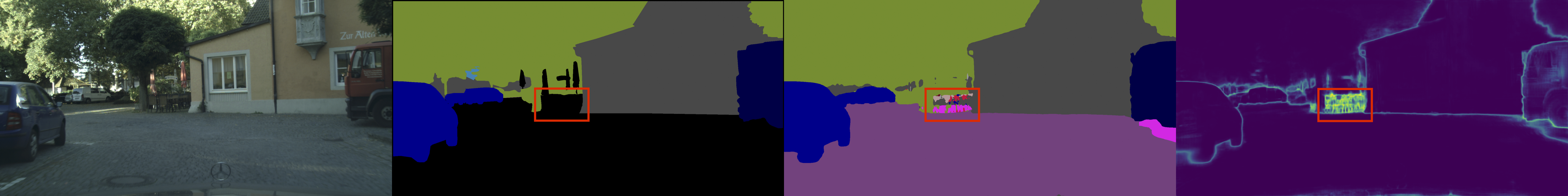}
        \caption{Supervised (ViT-B): Cityscapes}
    \end{subfigure}
    \vspace{0.2em}
    \begin{subfigure}{0.99\textwidth}
        \includegraphics[width=\textwidth]{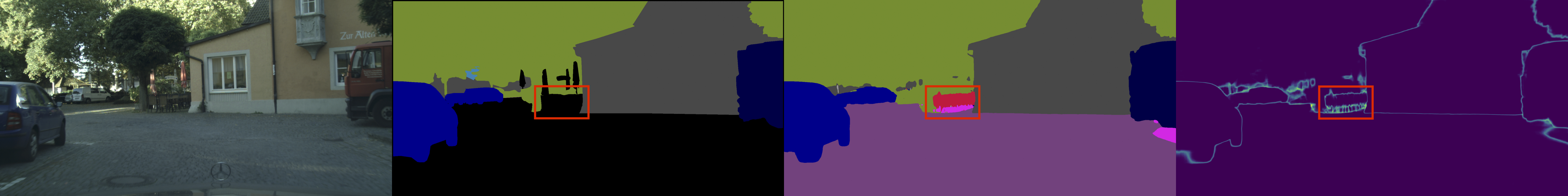}
        \caption{UniMatchV2 (ViT-B): Cityscapes}
    \end{subfigure}

    \begin{subfigure}{0.99\textwidth}
        \includegraphics[width=\textwidth]{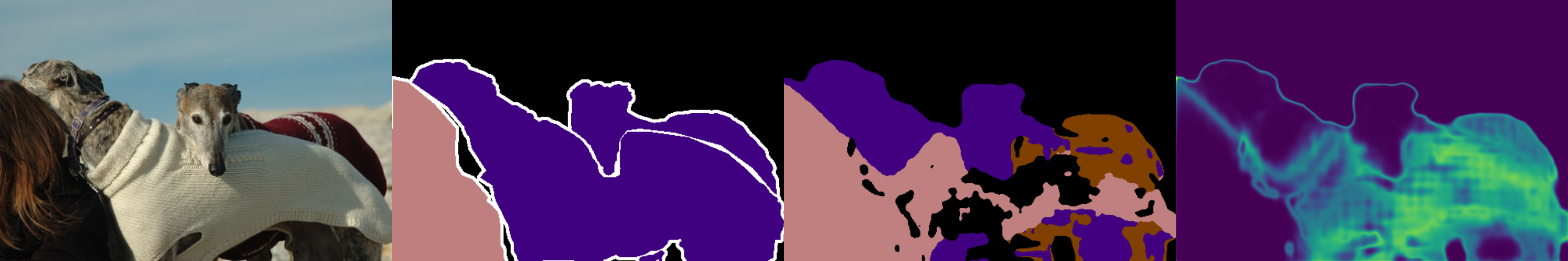}
        \caption{Supervised (ViT-B): Pascal VOC2012}
    \end{subfigure}
    \vspace{0.2em}
    \begin{subfigure}{0.99\textwidth}
        \includegraphics[width=\textwidth]{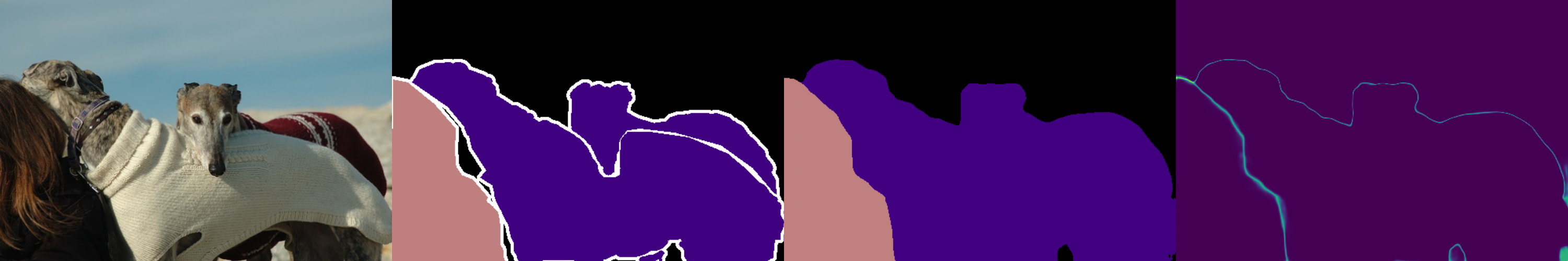}
        \caption{UniMatchV2 (ViT-B): Pascal VOC2012}
    \end{subfigure}

    \caption{Qualitative comparison between the supervised and UniMatchV2 models on Cityscapes (top) and Pascal VOC2012 (bottom). Each row shows from left to right: Input image, ground truth, prediction, and the corresponding uncertainty, computed using Shannon entropy (see Eq. \ref{eq: entropy}).}
    \label{fig:qualitative_comparison}
\end{figure*}

\textbf{Qualitative Analysis.}
Figure \ref{fig:qualitative_comparison} presents a qualitative comparison between the supervised baseline and the UniMatchV2 model on the Cityscapes and Pascal VOC2012 datasets. Models were trained on Cityscapes using only 1/8th of the labeled data, and on Pascal VOC2012 with just 92 labeled images. On Cityscapes, both models yield similar segmentation predictions overall. However, in a central image region lacking ground truth labels -- indicated by black pixels -- the supervised model produces noisy predictions but simultaneously yields high uncertainty, reflecting appropriate caution. In contrast, the UniMatchV2 model predicts multiple humans in this region with low uncertainty, raising concerns about overconfidence despite accurate-looking predictions.

For Pascal VOC2012, the supervised model fails to distinguish two dogs in the scene, while the UniMatchV2 model correctly segments both dogs with high confidence. Despite its errors, the supervised model flags the misclassified areas with high uncertainty, offering a valuable trade-off. Overall, this example highlights that while predictive accuracy is important, model reliability remains crucial for safe and trustworthy deployment, especially if we consider the impossibility of perfect segmentation performance -- especially in the case of out-of-domain scenarios, which we will examine next. 

\newpage

\subsection{Out-of-Domain Evaluation}
\begin{figure*}[!ht]
    \centering

    \begin{subfigure}{0.40\textwidth}
        \includegraphics[width=\textwidth]{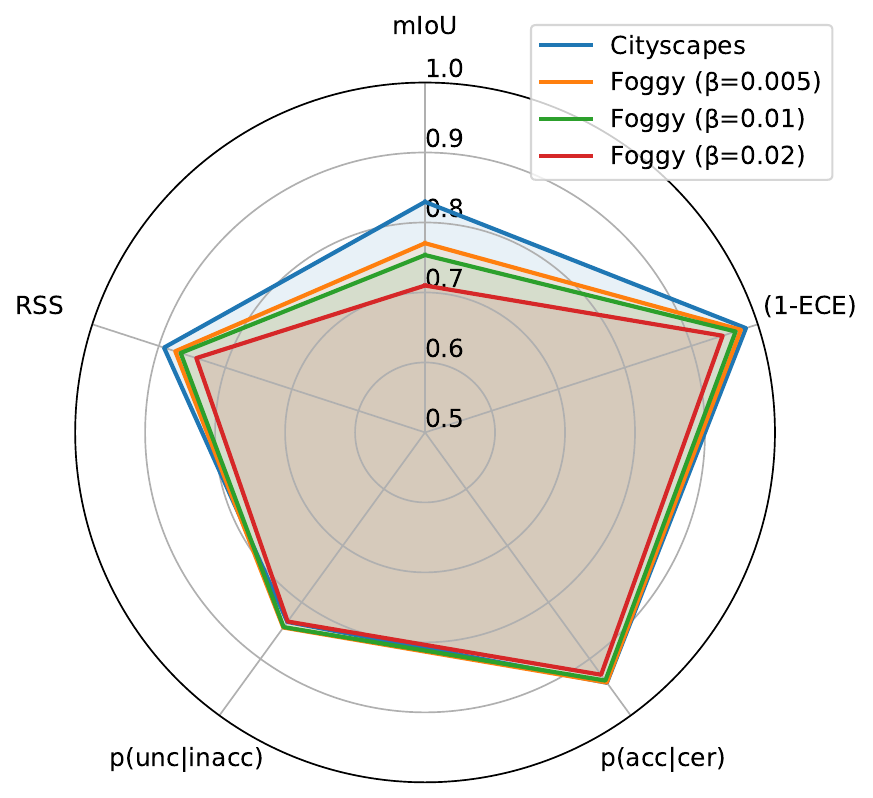}
        \caption{Supervised (ViT-B): Foggy}
    \end{subfigure}
    \vspace{0.2em}
    \begin{subfigure}{0.40\textwidth}
        \includegraphics[width=\textwidth]{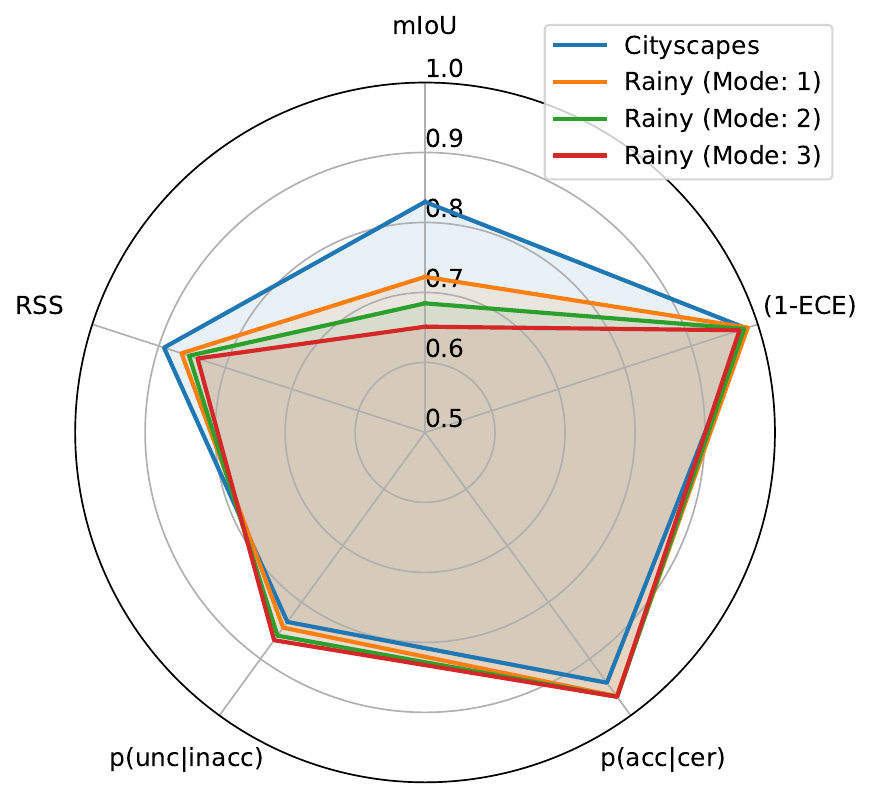}
        \caption{Supervised (ViT-B): Rainy}
    \end{subfigure}

    \begin{subfigure}{0.40\textwidth}
        \includegraphics[width=\textwidth]{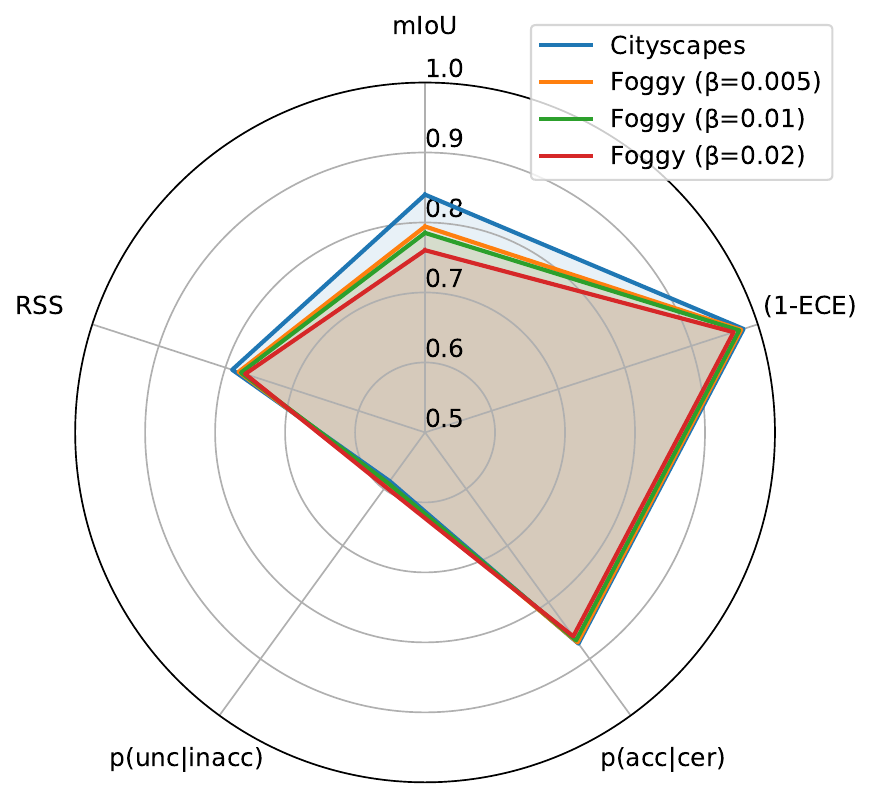}
        \caption{UniMatchV2 (ViT-B): Foggy}
    \end{subfigure}
    \vspace{0.2em}
    \begin{subfigure}{0.40\textwidth}
        \includegraphics[width=\textwidth]{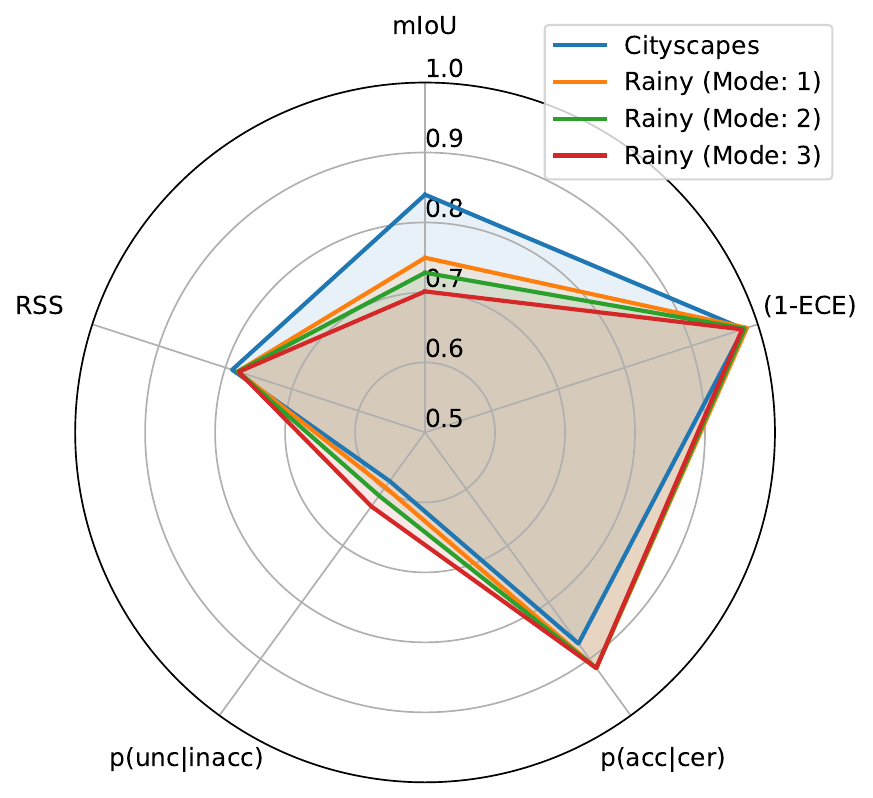}
        \caption{UniMatchV2 (ViT-B): Rainy}
    \end{subfigure}

    \caption{Quantitative comparison of the predictive performance (mIoU), calibration (ECE), and uncertainty (p(acc$|$cer), p(unc$|$inacc)) of the supervised baseline and UniMatchV2 under increasing levels of fog and rain on the Foggy and Rainy Cityscapes datasets.}
    \label{fig: ood_radar_analysis}
\end{figure*}

\textbf{Quantitative Analysis.} Figure \ref{fig: ood_radar_analysis} compares the robustness of the supervised baseline and UniMatchV2 on Foggy and Rainy Cityscapes datasets, using radar charts. Both models, trained on 1/8th of Cityscapes' labeled data, were evaluated in out-of-domain robustness without fine-tuning. As expected, segmentation performance degrades with increasing perturbation severity, but UniMatchV2 maintains higher mIoU values. Calibration similarly deteriorates under fog, albeit less than segmentation performance, especially for UniMatchV2, and remains comparatively stable under rainy conditions. Interestingly, uncertainty quality shows minimal change under fog and even improves under rain, particularly for UniMatchV2. Overall, UniMatchV2 exhibits enhanced robustness, with smaller RSS degradation due to improved uncertainty estimates despite declining accuracy. However, UniMatchV2 is ultimately still significantly less reliable due to its poor performance in terms of p(unc$|$inacc), barely exceeding 0.6 whereas the supervised baseline achieves values of approx. 0.85.

\begin{figure*}[!h]
    \centering

    \begin{subfigure}{0.99\textwidth}
        \includegraphics[width=\textwidth]{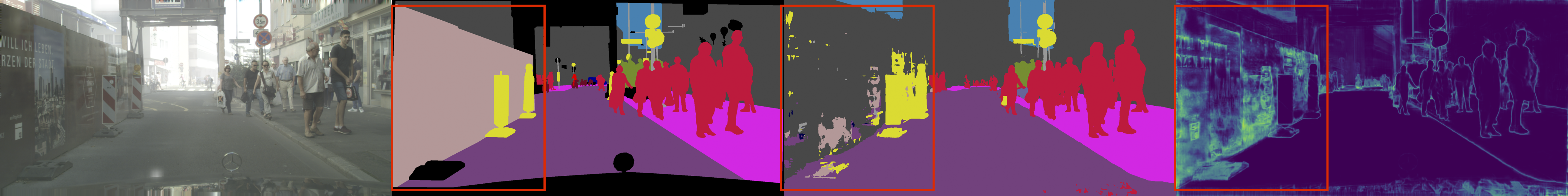}
        \caption{Supervised (ViT-B): Foggy}
    \end{subfigure}
    \vspace{0.2em}
    \begin{subfigure}{0.99\textwidth}
        \includegraphics[width=\textwidth]{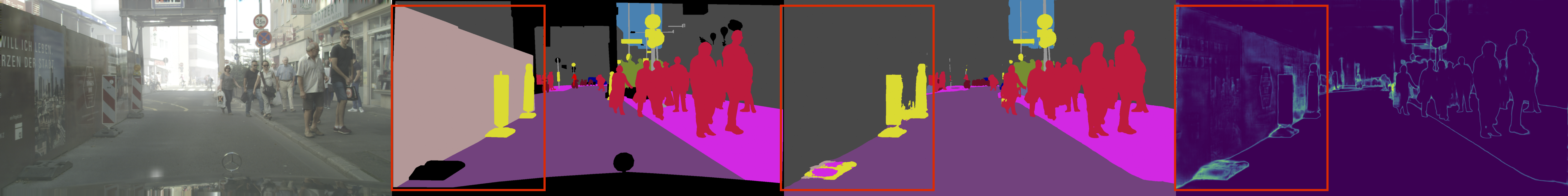}
        \caption{UniMatchV2 (ViT-B): Foggy}
    \end{subfigure}

    \begin{subfigure}{0.99\textwidth}
        \includegraphics[width=\textwidth]{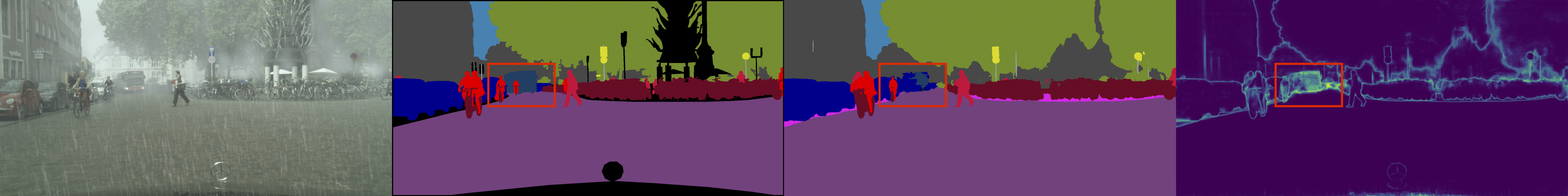}
        \caption{Supervised (ViT-B): Rainy}
    \end{subfigure}
    \vspace{0.2em}
    \begin{subfigure}{0.99\textwidth}
        \includegraphics[width=\textwidth]{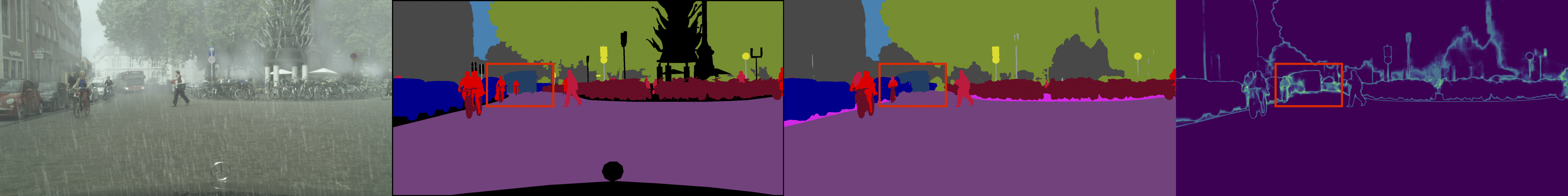}
        \caption{UniMatchV2 (ViT-B): Rainy}
    \end{subfigure}

    \caption{Qualitative comparison between the supervised and UniMatchV2 models on the strongest perturbed versions of Foggy (top) and Rainy Cityscapes (bottom). Each row shows from left to right: Input image, ground truth, prediction, and the corresponding uncertainty, computed using Shannon entropy (see Eq.~\ref{eq: entropy}).}
    \label{fig: qualitative_comparison_ood}
\end{figure*}

\textbf{Qualitative Analysis.} Figure \ref{fig: qualitative_comparison_ood} compares the supervised baseline and UniMatchV2 on the most perturbed versions of Foggy and Rainy Cityscapes. Both models were trained on 1/8 of the labeled data from the original Cityscapes dataset and not fine-tuned for this out-of-domain evaluation. In Foggy Cityscapes, both models misclassify the left region of the image as a building. However, while UniMatchV2 assigns low uncertainty to this incorrect prediction, the supervised model expresses high uncertainty, indicating an awareness of its error. In the Rainy Cityscapes example, UniMatchV2 yields slightly improved segmentation results, particularly in correctly identifying a bus in the background where the supervised model fails. Nonetheless, the supervised baseline assigns high uncertainty to the misclassified bus, in contrast to UniMatchV2's lower uncertainty. These examples highlight a trade-off between reliability and robustness, with UniMatchV2 achieving stronger performance but occasionally failing to reflect uncertainty when errors occur.


\section{Conclusion}
In this work, we addressed a critical blind spot in semi-supervised semantic segmentation: The lack of attention to model reliability and robustness. To fill this gap, we introduced the Reliable Segmentation Score (RSS), which holistically integrates accuracy, calibration, and uncertainty quality into a single metric through the harmonic mean, penalizing poor performance in any component. Through comprehensive evaluations on both in-domain and out-of-domain scenarios, we revealed that the current state-of-the-art UniMatchV2 achieves superior predictive performance and robustness but is often less calibrated and produces less reliable uncertainty estimates than its supervised counterpart. These findings raise legitimate questions about whether incremental gains in segmentation accuracy reflect meaningful progress toward reliable and robust deployment. We hope that our investigation and the proposed RSS metric serve as a stepping stone toward more principled evaluation protocols that better align research objectives with real-world requirements, focusing not only on performance but also on reliability and robustness.

%
%
%
%
\bibliographystyle{splncs04}
\bibliography{egbib}

\section*{Supplementary Material}
\begin{figure*}[!ht]
    \centering
    \begin{subfigure}{0.49\textwidth}
        \includegraphics[width=\textwidth]{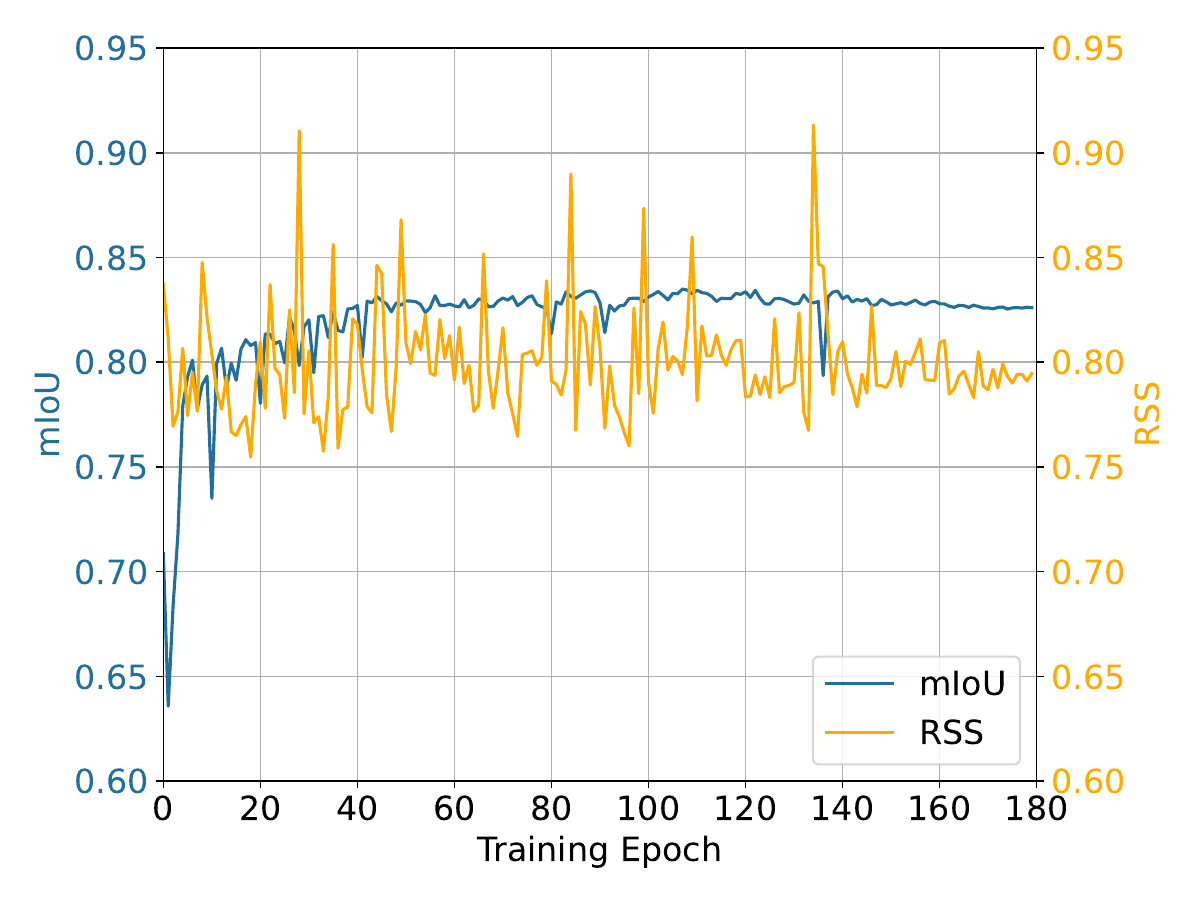}
        \caption{UniMatchV2: Cityscapes (1/16)}
        \label{subfig:miou_ece}
    \end{subfigure}
    \hfill
    \begin{subfigure}{0.49\textwidth}
        \includegraphics[width=\textwidth]{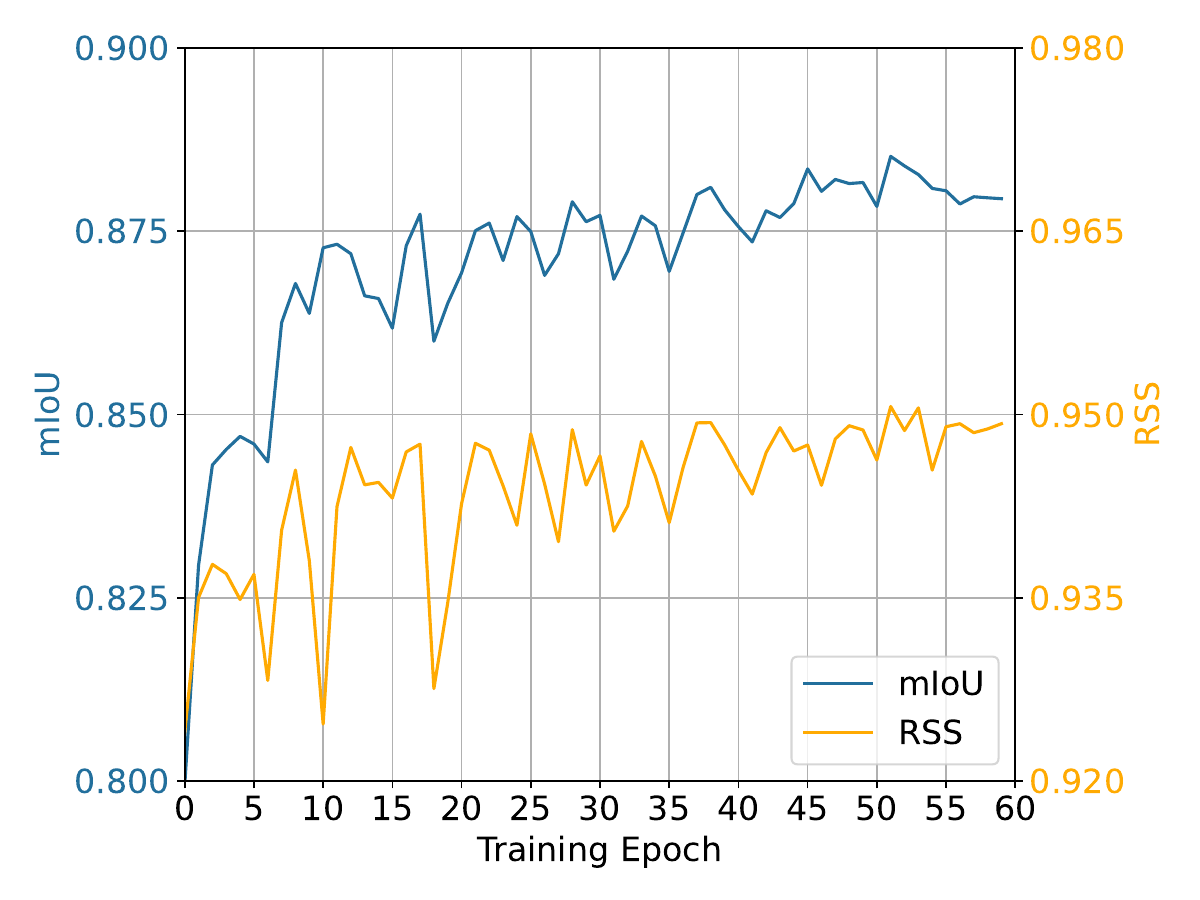}
        \caption{UniMatchV2: Pascal VOC2012 (183)}
        \label{subfig:uncertainty_metrics}
    \end{subfigure}
    \caption{Evolution of the mean Intersection over Union (mIoU) and Reliable Segmentation Score (RSS) metrics during UniMatchV2's training on the validation sets of Cityscapes and Pascal VOC2012.}
    \label{fig: mIoU_rss_training}
\end{figure*}

\textbf{Predictive Performance and Reliability during Training.} Figure \ref{fig: mIoU_rss_training} shows the evolution of mIoU and RSS on the validation sets of Cityscapes and Pascal VOC2012 during UniMatchV2 training. On Cityscapes, only 1/16th of the available labels were used, while for Pascal VOC2012, the model was trained with just 183 labeled images. For Cityscapes, the mIoU improves dramatically over time, starting below 0.65 after the first epoch and converging to around 0.84 between 80 and 120 training epochs. In contrast, the RSS fluctuates significantly across all epochs, reaching levels as high as 0.91 and as low as 0.76, but shows no consistent improvement over time. suggesting that the other three reliability metrics are even degrading during training since RSS incorporates mIoU in its calculation. On Pascal VOC2012, mIoU improves gradually from around 0.80 to 0.88, while the RSS also increases from approx. 0.92 to 0.95. However, this gain is less pronounced than that of the mIoU, again indicating that improvements in segmentation quality are not accompanied by improved reliability. These results reinforce our motivation to systematically investigate the reliability of semi-supervised segmentation methods, which are typically optimized solely for mIoU -- and apparently also at the expense of reliability. 

\newpage

\textbf{Quantitative Out-of-Domain Analysis.} Tables \ref{table: foggy_cityscapes} and \ref{table: rainy_cityscapes} corroborate the findings of the main part of the paper by comparing the segmentation accuracy, calibration, and uncertainty quality for all labeling fractions without fine-tuning on the most perturbed versions of Foggy and Rainy Cityscapes, respectively. Reminiscent of the finings in the main part, UniMatchV2 stands out as more robust, conveniently shown by a smaller degradation in RSS across all label fractions and both encoder sizes, but significantly less reliable due to its poor uncertainty quality. Overall, these results support our recommendation of rethinking semi-supervised segmentation evaluation protocols toward including model robustness alongside accuracy and reliability. 

\begin{table*}[!ht]
\begin{center}
\setlength\extrarowheight{1mm}
\newcolumntype{C}{>{\centering\arraybackslash}p{2.4cm}}
\resizebox{\textwidth}{!}{
\begin{tabular}{l|c|CCCCC}
\toprule
\multirow{2}{*}{} & \multirow{2}{*}{Encoder} 
& \multicolumn{5}{c}{Foggy Cityscapes (Label Fraction)} \\
\cmidrule(lr){3-7}
& & 1/30 & 1/16 & 1/8 & 1/4 & 1/2 \\ \hline \hline
\multicolumn{7}{c}{mIoU $\uparrow$} \\ \hline \hline
Supervised & ViT-S     & 0.573 (\textcolor{red}{-0.163}) & 0.623 (\textcolor{red}{-0.159}) & 0.653 (\textcolor{red}{-0.150}) & 0.679 (\textcolor{red}{-0.140}) & 0.680 (\textcolor{red}{-0.144}) \\
UniMatchV2 & ViT-S     & \textbf{0.691} (\textcolor{red}{\textbf{-0.102}}) & 0.700 (\textcolor{red}{-0.109}) & 0.697 (\textcolor{red}{-0.117}) & 0.706 (\textcolor{red}{-0.117}) & 0.716 (\textcolor{red}{-0.109}) \\
\cdashline{1-7}
Supervised & ViT-B     & 0.656 (\textcolor{red}{-0.132}) & 0.666 (\textcolor{red}{-0.145}) & 0.710 (\textcolor{red}{-0.119}) & 0.723 (\textcolor{red}{-0.117}) & 0.728 (\textcolor{red}{-0.118}) \\
UniMatchV2 & ViT-B     & 0.686 (\textcolor{red}{-0.121}) & \textbf{0.759} (\textcolor{red}{\textbf{-0.080}}) & \textbf{0.760} (\textcolor{red}{\textbf{-0.080}}) & \textbf{0.754} (\textcolor{red}{\textbf{-0.093}}) & \textbf{0.758} (\textcolor{red}{\textbf{-0.090}}) \\ \hline \hline
\multicolumn{7}{c}{ECE $\downarrow$} \\ \hline \hline
Supervised & ViT-S     & 0.063 (\textcolor{red}{0.047}) & 0.081 (\textcolor{red}{0.058}) & 0.061 (\textcolor{red}{0.045}) & 0.047 (\textcolor{red}{0.034}) & 0.036 (\textcolor{red}{\textbf{0.000}}) \\
UniMatchV2 & ViT-S     & 0.057 (\textcolor{red}{\textbf{0.028}}) & 0.059 (\textcolor{red}{0.032}) & 0.054 (\textcolor{red}{0.032}) & 0.051 (\textcolor{red}{0.033}) & 0.040 (\textcolor{red}{0.029}) \\
\cdashline{1-7}
Supervised & ViT-B     & 0.060 (\textcolor{red}{0.040}) & 0.084 (\textcolor{red}{0.059}) & 0.053 (\textcolor{red}{0.035}) & 0.046 (\textcolor{red}{0.031}) & 0.040 (\textcolor{red}{0.029}) \\
UniMatchV2 & ViT-B     & \textbf{0.049} (\textcolor{red}{0.030}) & \textbf{0.039} (\textcolor{red}{\textbf{0.014}}) & \textbf{0.036} (\textcolor{red}{\textbf{0.014}}) & \textbf{0.037} (\textcolor{red}{\textbf{0.022}}) & \textbf{0.031} (\textcolor{red}{0.020}) \\ \hline \hline
\multicolumn{7}{c}{p(acc$|$cer) $\uparrow$} \\ \hline \hline
Supervised & ViT-S     & 0.868 (\textcolor{red}{-0.012}) & 0.884 (\textcolor{red}{-0.016}) & 0.874 (\textcolor{red}{\textbf{-0.002}}) & 0.885 (\textcolor{ForestGreen}{\textbf{0.005}}) & \textbf{0.910} (\textcolor{ForestGreen}{\textbf{0.003}}) \\
UniMatchV2 & ViT-S     & 0.853 (\textcolor{red}{-0.019}) & 0.856 (\textcolor{red}{\textbf{-0.015}}) & 0.859 (\textcolor{red}{-0.003}) & 0.860 (\textcolor{red}{-0.005}) & 0.853 (\textcolor{red}{-0.001}) \\
\cdashline{1-7}
Supervised & ViT-B     & \textbf{0.925} (\textcolor{red}{\textbf{-0.001}}) & \textbf{0.924} (\textcolor{red}{-0.018}) & \textbf{0.928} (\textcolor{red}{-0.014}) & \textbf{0.907} (\textcolor{red}{-0.001}) & 0.897 (\textcolor{red}{-0.008}) \\
UniMatchV2 & ViT-B     & 0.884 (\textcolor{red}{-0.006}) & 0.850 (\textcolor{red}{-0.020}) & 0.860 (\textcolor{red}{-0.013}) & 0.850 (\textcolor{red}{-0.007}) & 0.849 (\textcolor{red}{-0.003}) \\ \hline \hline
\multicolumn{7}{c}{p(unc$|$inacc) $\uparrow$} \\ \hline \hline
Supervised & ViT-S     & 0.731 (\textcolor{ForestGreen}{\textbf{0.068}}) & 0.759 (\textcolor{ForestGreen}{\textbf{0.039}}) & 0.717 (\textcolor{ForestGreen}{\textbf{0.076}}) & 0.732 (\textcolor{ForestGreen}{\textbf{0.084}}) & \textbf{0.780} (\textcolor{ForestGreen}{0.067}) \\
UniMatchV2 & ViT-S     & 0.624 (\textcolor{ForestGreen}{0.018}) & 0.639 (\textcolor{ForestGreen}{0.033}) & 0.640 (\textcolor{ForestGreen}{0.075}) & 0.641 (\textcolor{ForestGreen}{0.065}) & 0.611 (\textcolor{ForestGreen}{\textbf{0.081}}) \\
\cdashline{1-7}
Supervised & ViT-B     & \textbf{0.840} (\textcolor{ForestGreen}{0.046}) & \textbf{0.852} (\textcolor{ForestGreen}{0.010}) & \textbf{0.834} (\textcolor{red}{-0.001}) & \textbf{0.760} (\textcolor{ForestGreen}{0.051}) & 0.736 (\textcolor{ForestGreen}{0.038}) \\
UniMatchV2 & ViT-B     & 0.721 (\textcolor{ForestGreen}{0.048}) & 0.582 (\textcolor{red}{-0.004}) & 0.596 (\textcolor{ForestGreen}{0.010}) & 0.580 (\textcolor{ForestGreen}{0.047}) & 0.578 (\textcolor{ForestGreen}{0.063}) \\ \hline \hline

\multicolumn{7}{c}{RSS $\uparrow$} \\ \hline \hline
Supervised & ViT-S & 0.750 (\textcolor{red}{-0.047}) & 0.778 (\textcolor{red}{-0.055}) & 0.779 (\textcolor{red}{-0.027}) & 0.797 (\textcolor{red}{-0.017}) & \textbf{0.818} (\textcolor{red}{-0.023}) \\
UniMatchV2 & ViT-S & 0.757 (\textcolor{red}{-0.029}) & 0.766 (\textcolor{red}{\textbf{-0.024}}) & 0.767 (\textcolor{red}{\textbf{-0.005}}) & 0.771 (\textcolor{red}{-0.009}) & 0.762 (\textcolor{ForestGreen}{0.004}) \\
\cdashline{1-7}
Supervised & ViT-B & \textbf{0.823} (\textcolor{red}{-0.041}) & \textbf{0.825} (\textcolor{red}{-0.063}) & \textbf{0.844} (\textcolor{red}{-0.048}) & \textbf{0.825} (\textcolor{red}{-0.023}) & \textbf{0.818} (\textcolor{red}{-0.027}) \\
UniMatchV2 & ViT-B & 0.795 (\textcolor{red}{\textbf{-0.027}}) & 0.762 (\textcolor{red}{-0.027}) & 0.770 (\textcolor{red}{-0.020}) & 0.760 (\textcolor{red}{\textbf{-0.003}}) & 0.761 (\textcolor{ForestGreen}{\textbf{0.007}}) \\ \hline \hline
\end{tabular}}
\end{center}
\caption{Quantitative performance, calibration, and uncertainty robustness evaluation without fine-tuning on the strongest perturbation setting on Foggy Cityscapes ($\beta = 0.02$). Values in brackets highlight differences to the original Cityscapes results.}
\label{table: foggy_cityscapes}
\end{table*}

\begin{table*}[!ht]
\begin{center}
\setlength\extrarowheight{1mm}
\newcolumntype{C}{>{\centering\arraybackslash}p{2.4cm}}
\resizebox{\textwidth}{!}{
\begin{tabular}{l|c|CCCCC}
\toprule
\multirow{2}{*}{} & \multirow{2}{*}{Encoder} 
& \multicolumn{5}{c}{Rainy Cityscapes (Label Fraction)} \\
\cmidrule(lr){3-7}
& & 1/30 & 1/16 & 1/8 & 1/4 & 1/2 \\ \hline \hline
\multicolumn{7}{c}{mIoU $\uparrow$} \\ \hline \hline
Supervised & ViT-S     & 0.520 (\textcolor{red}{-0.216}) & 0.572 (\textcolor{red}{-0.210}) & 0.611 (\textcolor{red}{-0.192}) & 0.595 (\textcolor{red}{-0.224}) & 0.658 (\textcolor{red}{-0.166}) \\
UniMatchV2 & ViT-S     & \textbf{0.616} (\textcolor{red}{\textbf{-0.177}}) & 0.606 (\textcolor{red}{-0.203}) & 0.633 (\textcolor{red}{-0.181}) & 0.654 (\textcolor{red}{-0.169}) & 0.651 (\textcolor{red}{-0.174}) \\
\cdashline{1-7}
Supervised & ViT-B     & 0.578 (\textcolor{red}{-0.210}) & 0.625 (\textcolor{red}{-0.186}) & 0.651 (\textcolor{red}{-0.178}) & 0.661 (\textcolor{red}{-0.179}) & 0.652 (\textcolor{red}{-0.194}) \\
UniMatchV2 & ViT-B     & 0.608 (\textcolor{red}{-0.199}) & \textbf{0.682} (\textcolor{red}{\textbf{-0.157}}) & \textbf{0.701} (\textcolor{red}{\textbf{-0.139}}) & \textbf{0.697} (\textcolor{red}{\textbf{-0.150}}) & \textbf{0.698} (\textcolor{red}{\textbf{-0.150}}) \\ \hline \hline
\multicolumn{7}{c}{ECE $\downarrow$} \\ \hline \hline
Supervised & ViT-S     & 0.038 (\textcolor{red}{0.022}) & 0.044 (\textcolor{red}{0.021}) & 0.030 (\textcolor{red}{0.014}) & 0.021 (\textcolor{red}{0.008}) & 0.031 (\textcolor{ForestGreen}{\textbf{-0.005}}) \\
UniMatchV2 & ViT-S     & 0.038 (\textcolor{red}{0.009}) & 0.042 (\textcolor{red}{0.015}) & 0.031 (\textcolor{red}{0.009}) & 0.024 (\textcolor{red}{0.006}) & 0.018 (\textcolor{red}{0.007}) \\
\cdashline{1-7}
Supervised & ViT-B     & 0.030 (\textcolor{red}{0.010}) & 0.045 (\textcolor{red}{0.020}) & 0.028 (\textcolor{red}{0.010}) & 0.022 (\textcolor{red}{0.007}) & 0.017 (\textcolor{red}{0.006}) \\
UniMatchV2 & ViT-B     & \textbf{0.025} (\textcolor{red}{\textbf{0.006}}) & \textbf{0.028} (\textcolor{red}{\textbf{0.003}}) & \textbf{0.024} (\textcolor{red}{\textbf{0.002}}) & \textbf{0.020} (\textcolor{red}{\textbf{0.005}}) & \textbf{0.013} (\textcolor{red}{0.002}) \\ \hline \hline
\multicolumn{7}{c}{p(acc$|$cer) $\uparrow$} \\ \hline \hline
Supervised & ViT-S     & 0.918 (\textcolor{ForestGreen}{0.038}) & 0.940 (\textcolor{ForestGreen}{0.040}) & 0.936 (\textcolor{ForestGreen}{\textbf{0.060}}) & 0.926 (\textcolor{ForestGreen}{0.046}) & \textbf{0.921} (\textcolor{ForestGreen}{0.014}) \\
UniMatchV2 & ViT-S     & 0.915 (\textcolor{ForestGreen}{0.043}) & 0.922 (\textcolor{ForestGreen}{\textbf{0.051}}) & 0.918 (\textcolor{ForestGreen}{0.056}) & 0.915 (\textcolor{ForestGreen}{0.050}) & 0.912 (\textcolor{ForestGreen}{\textbf{0.058}}) \\
\cdashline{1-7}
Supervised & ViT-B     & \textbf{0.959} (\textcolor{ForestGreen}{0.033}) & \textbf{0.962} (\textcolor{ForestGreen}{0.020}) & \textbf{0.967} (\textcolor{ForestGreen}{0.025}) & \textbf{0.947} (\textcolor{ForestGreen}{0.039}) & 0.914 (\textcolor{ForestGreen}{0.009}) \\
UniMatchV2 & ViT-B     & 0.939 (\textcolor{ForestGreen}{\textbf{0.049}}) & 0.915 (\textcolor{ForestGreen}{0.045}) & 0.916 (\textcolor{ForestGreen}{0.043}) & 0.909 (\textcolor{ForestGreen}{\textbf{0.052}}) & 0.908 (\textcolor{ForestGreen}{0.056}) \\ \hline \hline
\multicolumn{7}{c}{p(unc$|$inacc) $\uparrow$} \\ \hline \hline
Supervised & ViT-S     & 0.741 (\textcolor{ForestGreen}{0.078}) & 0.792 (\textcolor{ForestGreen}{0.072}) & 0.757 (\textcolor{ForestGreen}{\textbf{0.116}}) & 0.707 (\textcolor{ForestGreen}{0.059}) & \textbf{0.686} (\textcolor{red}{-0.027}) \\
UniMatchV2 & ViT-S     & 0.672 (\textcolor{ForestGreen}{0.066}) & 0.702 (\textcolor{ForestGreen}{\textbf{0.096}}) & 0.670 (\textcolor{ForestGreen}{0.105}) & 0.649 (\textcolor{ForestGreen}{0.073}) & 0.634 (\textcolor{ForestGreen}{\textbf{0.104}}) \\
\cdashline{1-7}
Supervised & ViT-B     & \textbf{0.843} (\textcolor{ForestGreen}{0.049}) & \textbf{0.862} (\textcolor{ForestGreen}{0.020}) & \textbf{0.867} (\textcolor{ForestGreen}{0.032}) & \textbf{0.783} (\textcolor{ForestGreen}{\textbf{0.074}}) & 0.649 (\textcolor{red}{-0.049}) \\
UniMatchV2 & ViT-B     & 0.757 (\textcolor{ForestGreen}{\textbf{0.084}}) & 0.628 (\textcolor{ForestGreen}{0.042}) & 0.631 (\textcolor{ForestGreen}{0.045}) & 0.601 (\textcolor{ForestGreen}{0.068}) & 0.600 (\textcolor{ForestGreen}{0.085}) \\ \hline \hline
\multicolumn{7}{c}{RSS $\uparrow$} \\ \hline \hline
Supervised & ViT-S & 0.740 (\textcolor{red}{-0.057}) & 0.781 (\textcolor{red}{-0.052}) & 0.791 (\textcolor{red}{-0.015}) & 0.770 (\textcolor{red}{-0.044}) & \textbf{0.785} (\textcolor{red}{-0.056}) \\
UniMatchV2 & ViT-S & 0.763 (\textcolor{red}{\textbf{-0.023}}) & 0.769 (\textcolor{red}{-0.021}) & 0.770 (\textcolor{red}{\textbf{-0.002}}) & 0.771 (\textcolor{red}{-0.009}) & 0.765 (\textcolor{ForestGreen}{\textbf{0.007}}) \\
\cdashline{1-7}
Supervised & ViT-B & \textbf{0.802} (\textcolor{red}{-0.062}) & \textbf{0.825} (\textcolor{red}{-0.063}) & \textbf{0.842} (\textcolor{red}{-0.050}) & \textbf{0.822} (\textcolor{red}{-0.026}) & 0.771 (\textcolor{red}{-0.074}) \\
UniMatchV2 & ViT-B & 0.791 (\textcolor{red}{-0.031}) & 0.772 (\textcolor{red}{\textbf{-0.017}}) & 0.780 (\textcolor{red}{-0.010}) & 0.767 (\textcolor{ForestGreen}{\textbf{0.004}}) & 0.767 (\textcolor{ForestGreen}{0.013}) \\ \bottomrule
\end{tabular}}
\end{center}
\caption{Quantitative performance, calibration, and uncertainty robustness evaluation without fine-tuning on the strongest perturbation setting on Rainy$_3$, which uses [0.03, 0.015, 0.002] as attenuation coefficients $\alpha$ and $\beta$ and the raindrop radius $a$. $\alpha$ and $\beta$ determine the degree of simulated rain and fog in the images. Values in brackets highlight differences to the original Cityscapes results.}
\label{table: rainy_cityscapes}
\end{table*}

\end{document}